\documentclass[10pt]{article}
\usepackage[preprint]{tmlr}
\usepackage{algorithm}
\let\classAND\AND
\let\AND\relax
\usepackage{algorithmic}

\let\AND\classAND
\AtBeginEnvironment{algorithmic}{\let\AND\algoAND}
\usepackage{amsthm,amsmath,amssymb,eucal}
\usepackage{graphicx, color}
\usepackage[T1]{fontenc}
\usepackage{pgffor}

\usepackage{float}
\usepackage{xcolor}
\usepackage[labelfont=bf]{caption}
\usepackage{subcaption}
\usepackage[hidelinks]{hyperref}
\usepackage{makecell}
\usepackage{comment}
\usepackage{svg}
\usepackage{multirow}
\usepackage{booktabs}

\title{Improving Performance in Classification Tasks with LCEN and the Weighted Focal Differentiable MCC Loss}

\author{
\name Pedro Seber \email pseber@mit.edu \\
\addr Department of Chemical Engineering \\
Massachusetts Institute of Technology
\AND
\name Richard D.\ Braatz \email braatz@mit.edu \\
\addr Department of Chemical Engineering \\
Massachusetts Institute of Technology
}

\begin{document}
\maketitle
\begin{abstract}
The LASSO-Clip-EN (LCEN) algorithm was previously introduced for nonlinear, interpretable feature selection and machine learning. However, its design and use was limited to regression tasks. In this work, we create a modified version of the LCEN algorithm that is suitable for classification tasks and maintains its desirable properties, such as interpretability. This modified LCEN algorithm is evaluated on four widely used binary and multiclass classification datasets. In these experiments, LCEN is compared against 10 other model types and consistently reaches high test-set macro F\(_1\) score and Matthews correlation coefficient (MCC) metrics, higher than that of the majority of investigated models. LCEN models for classification remain sparse, eliminating an average of 56\% of all input features in the experiments performed. Furthermore, LCEN-selected features are used to retrain all models using the same data, leading to statistically significant performance improvements in three of the experiments and insignificant differences in the fourth when compared to using all features or other feature selection methods. Simultaneously, the weighted focal differentiable MCC (diffMCC) loss function is evaluated on the same datasets. Models trained with the diffMCC loss function are always the best-performing methods in these experiments, and reach test-set macro F\(_1\) scores that are, on average, 4.9\% higher and MCCs that are 8.5\% higher than those obtained by models trained with the weighted cross-entropy loss. These results highlight the performance of LCEN as a feature selection and machine learning algorithm also for classification tasks, and how the diffMCC loss function can train very accurate models, surpassing the weighted cross-entropy loss in the tasks investigated.
\end{abstract}

\section{Introduction}
Classification problems are ubiquitous in the sciences, engineering, and other important areas. Many machine learning models for classification exist, each with different capabilities and desirable properties. Examples of desirable properties include interpretability (as defined by \citet{Seber-and-Braatz-2025-LCEN}, ``how an output $y = f(X)$ was predicted for a given input $X$ --- that is, provide $f(\cdot)$ in a form readily understandable to humans'') and sparsity (``a model that uses few input features, particularly relative to the total number of features available.'' \citep{Seber-and-Braatz-2025-LCEN}). Interpretability is an essential property for a model to be trusted in critical or sensitive applications \citep{Hong-etal-2020}, which are widely found in many fields. If an interpretable model is not used, interpretability may be obtained indirectly through \textit{a posteriori} methods, such as Shapley values \citep{Shapley-1951} or LIME \citep{Ribeiro-etal-2016}. These methods can be useful and increase the trustworthiness of models, but previous works have found that they are unreliable \citep{Rudin-2019}, a problem exacerbated by ``shortcuts'' taken by some models \citep{Lapuschkin-etal-2019,Rosenzweig-etal-2021}. These ``shortcuts'' apparently increase predictive capabilities but are unrelated to the actual problem, leading to issues with predictive performance generalization.

Feature selection is one technique to improve the sparsity of models, which, in turn, can improve interpretability to humans if the method is interpretable. Feature selection involves determining the most important features among those available. There are many methods for feature selection, which can be broadly classified into wrapper, filter, and embedded methods. A summary of feature selection methods for regression tasks, some of which also function for classification, can be found in the Introduction section of \citet{Seber-and-Braatz-2025-LCEN}. The works of \citet{Dasgupta-etal-2007}, \citet{Jovic-etal-2015}, and \citet{Jiao-etal-2024} provide comprehensive reviews on feature selection methods for different classification tasks (including, but not limited to, text mining or image-based classification).

Recently, the LASSO-Clip-EN (LCEN) algorithm has been introduced for nonlinear, interpretable feature selection and machine learning in regression tasks \citep{Seber-and-Braatz-2025-LCEN}. LCEN had strong feature selection capabilities (comparable to that of thresholded EN, but LCEN was 10.3-fold faster) and machine learning performance (LCEN had low test RMSEs in multiple tested datasets and was among the best methods tested) \citep{Seber-and-Braatz-2025-LCEN}. However, an important limitation of \citet{Seber-and-Braatz-2025-LCEN} (noted in that work) is that it considered only regression tasks. Nevertheless, an analogous algorithm can be created and used for nonlinear, interpretable feature selection and machine learning in classification tasks.

In opposition to what has been described above, there are also scenarios in which the priority is the maximization or minimization of a performance metric or metrics (such as the mean squared error (MSE), F\(_1\) score, or precision), even at the cost of interpretability or other desirable properties. Artificial neural networks (ANNs) tend to dominate these scenarios due to their high potential for accurate modeling, in part because ANNs are universal function approximators \citep{Cybenko-1989,Hornik-etal-1989,Hornik-1991}. As noted by \citet{Seber-2024}, in classification tasks, there is a discrepancy between what is optimized and the classification metrics. Deep learning models for classification are typically trained with the (weighted) cross-entropy (CE) loss, which aims to minimize the cross-entropy between the model's predictions and the labels from data. However, classification models are then evaluated based on other classification metrics such as the F\(_1\) score or the Matthews correlation coefficient (MCC). This discrepancy between the loss function and the target metric can mean that the model with the lowest cross-entropy is not necessarily the model with the best metrics. Classification metrics are typically not differentiable, so they cannot be directly used as a loss function, but a differentiable variant of the MCC was used by \citet{Seber-2024} to create the weighted focal differentiable MCC loss function (diffMCC). In contrast, this issue does not occur in regression tasks because the loss function and evaluation metric are typically the MSE, and most evaluation metrics are a function of \(y - \hat{y}\), which is optimized by the MSE loss.

In this work, a modification of the LCEN algorithm for classification tasks is also described and evaluated. This modified algorithm is tested on a wide range of classification problems, including binary and multiclass tasks related to the physical, biological, and social sciences. As done in \citet{Seber-and-Braatz-2025-LCEN}, ablated and variant LCEN algorithms for classification are tested to show that LCEN, and not other algorithms, is also optimal for the tasks investigated. In the four real datasets tested in this work, LCEN reaches higher macro F\(_1\) scores and MCCs than the majority of machine learning models evaluated; its performance is only 7.3\% lower in macro F\(_1\) score and 15.0\% lower in MCC than the best model of each dataset. LCEN is consistently among the top models in the datasets tested, something only MLPs can also achieve, and LCEN can even surpass these deep learning models in some cases. This high performance occurs at no cost to LCEN's interpretability or sparsity. LCEN consistently created sparse models, eliminating an average of 56\% of all input features and more than 50\% of the input features on 3/4 of the tested datasets. Furthermore, LCEN's feature selection capabilities are also tested by retraining every model on a dataset containing only features selected by LCEN. The average performance (in terms of macro F\(_1\) scores and MCCs) of all machine learning models tested is improved in 3/4 of the tested datasets and unchanged in the last one when compared to the results obtained with all features or with features selected by previous works through different methods. Statistical analysis of the performance improvements is done with Tukey's HSD test \citep{Tukey-1949}. These results demonstrate that LCEN can select the most important features in a given dataset, allowing models to be improved by reducing their data variance. As previously mentioned by \citet{Seber-and-Braatz-2025-LCEN}, the high performance of LCEN has a theoretical basis thanks to ``important previous works that proved desirable theoretical properties of the thresholded LASSO (a LASSO-Clip model) and the thresholded EN (an EN-Clip model)'' \citep{Zhou-2009,Meinshausen-and-Yu-2009,Zou-and-Zhang-2009,Zhou-2010,van-de-Geer-etal-2011}. 

Finally, we also systematically evaluate the diffMCC loss function on these same datasets and compare it to the weighted cross-entropy loss function. In these four experiments, the diffMCC loss always trained MLP models with higher test-set performance than those trained with the cross-entropy loss. On average across all experiments in this work, MLPs trained with the diffMCC loss function reach 4.9\% higher test-set macro F\(_1\) scores and 8.5\% higher MCCs than those achieved by MLPs trained with the weighted CE.

This work is organized as follows: first, the modifications done to the LCEN algorithm so that it can also work in classification tasks are described in Section \ref{Section_diffMCC_LCEN_classification_modifying_LCEN}; these are followed by comments on the diffMCC loss function and the experimental setup done in this work (Section \ref{Section_diffMCC_LCEN_classification_setup}). Section \ref{Section_diffMCC_LCEN_classification_results_binary} provides results on two real binary classification datasets (Tables \ref{Table_diffMCC_LCEN_classification_results_heart_failure} and \ref{Table_diffMCC_LCEN_classification_results_bank_marketing}). Section \ref{Section_diffMCC_LCEN_classification_results_clip_min_class} investigates the feature-selection effects of a new hyperparameter used in the modified LCEN algorithm for classification (Table \ref{Table_diffMCC_LCEN_classification_results_LCEN_clip_FS}); this hyperparameter was introduced in point 2 of Section \ref{Section_diffMCC_LCEN_classification_modifying_LCEN}. Section \ref{Section_diffMCC_LCEN_classification_results_multiclass} provides results on two real multiclass classification datasets (Tables \ref{Table_diffMCC_LCEN_classification_results_wine} and \ref{Table_diffMCC_LCEN_classification_results_glass}). Finally, these results and future directions are discussed in Section \ref{Section_diffMCC_LCEN_classification_discussion}.

\section{Methods}
\subsection{Modifying the LCEN algorithm for classification} \label{Section_diffMCC_LCEN_classification_modifying_LCEN}
The LCEN algorithm (Algorithm 1 of \citet{Seber-and-Braatz-2025-LCEN}) was originally developed only for regression tasks. To create an analogous version for classification tasks, we make the following modifications to the algorithm:
\begin{enumerate}
\item The LASSO step, which involves linear regression with an L\(_1\) norm penalty, is modified to use logistic regression with an L\(_1\) norm penalty.
\item The clip steps remain unchanged for binary classification scenarios. In multiclass classification, one set of features is generated for each class; thus, they need to be harmonized before the EN step or final output. Within the algorithm, the hyperparameter \texttt{LCEN\_clip\_min\_classes\_selected} controls the minimum number of feature sets in which a feature must appear to not be eliminated by the clip step. For example, a feature that is selected\footnote{``Selected'' follows the definition from the clip step of the original LCEN algorithm, in which a feature is selected if its scaled coefficient is \(\ge\) the \texttt{cutoff} hyperparameter.} by 2 or fewer classes will be discarded if the \texttt{LCEN\_clip\_min\_classes\_selected} is \(\ge 3\).
\item The EN step, which involves linear regression with a combined L\(_1\)-L\(_2\) norm penalty, is modified to use logistic regression with a combined L\(_1\)-L\(_2\) norm penalty.
\end{enumerate}

These modifications maintain the same optimal strategies and properties mentioned in Sections 2.1 and 2.2 of \citet{Seber-and-Braatz-2025-LCEN}, but allow LCEN to also be used as a feature selection and machine learning algorithm for classification problems. Ablation tests are performed in Section \ref{Section_diffMCC_LCEN_classification_appendix_ablation} to confirm the superior performance of LCEN over ablated and variant algorithms, also in the classification tasks evaluated. In all datasets investigated, LCEN reached the highest macro F\(_1\) score and MCC metrics (tied in one case with LC, and by itself in all others) (Tables \ref{Table_diffMCC_LCEN_classification_ablation_heart_failure}--\ref{Table_diffMCC_LCEN_classification_ablation_glass}). Unlike in \citet{Seber-and-Braatz-2025-LCEN}, no specific patterns arose among the results from ablated and variant models. For example, EN-Clip (ENC) tended to have lower performances, and LASSO-Clip (LC) was the worst model on one dataset and tied with LCEN for best model on another. Nevertheless, LCEN was the only model that displayed consistently high performance in these experiments, corroborating the notion that it is the best model among all variants and ablations. 

The classification form of the LCEN algorithm uses the same hyperparameters and has the same scaling as the regression form. With \(N\) as the number of samples in a dataset, \(P\) as the number of features in a dataset after feature expansion (Algorithm 2 of \citet{Seber-and-Braatz-2025-LCEN}), \(F\) as the number of cross-validation folds, \(A\) as the number of regularization strength (\(\alpha\)) values tested, \(C\) as the number of cutoffs tested, \(p\) as the number of features remaining after the LASSO and Clip steps, and \(L\) as the number of L\(_1\) ratios tested, LCEN scales as O(\(NP^2FAC + Np^2FALC\)). Although empirical measurements of runtime were not performed in this work, we introduce a brief proof showing when LCEN is faster than ENC:
\begin{align*}
t_{\text{LCEN}} = \text{O}(NP^2FAC + Np^2FALC) \text{ and } t_{\text{ENC}} = \text{O}(NP^2FALC)\\
\\
t_{\text{LCEN}} < t_{\text{ENC}}  \rightarrow \\
NP^2FAC + Np^2FALC < NP^2FALC \rightarrow \\
P^2 + p^2L < P^2L \rightarrow \\
p^2L < P^2(L-1) \rightarrow \\
\boxed{p < P\sqrt{\frac{L-1}{L}}}
\end{align*}
In this work and in \citet{Seber-and-Braatz-2025-LCEN}, \(L = 13\), so \(p < 0.96P\) for \(t_{\text{LCEN}} < t_{\text{ENC}}\). Only 4\% of the input features need to be eliminated by the LASSO and Clip steps for LCEN to be faster than ENC, an almost-guaranteed phenomenon. As ENCEN is slower than ENC, LCEN is faster than ENCEN as well.

\subsection{Other classification methods and experimental setup} \label{Section_diffMCC_LCEN_classification_setup}
In addition to LCEN, other algorithms are compared in the experiments of this work, including models from previously published works. In particular, this work also focuses on ANNs trained with the weighted focal differentiable MCC (diffMCC) loss function (Algorithms 1 and 2 of \citet{Seber-2024}). This loss function was created to address shortcomings with the (weighted) cross-entropy loss, the most used loss function in classification tasks. The diffMCC loss function was claimed to outperform the weighted cross-entropy \citep{Seber-2024}. However, this loss function has been used only in a few problems of biological relevance \citep{Seber-2024,Seber-and-Braatz-2025}, so there is a need for an analysis of its potential performance improvements over a wider series of problems, including those unrelated to biology.

As in \citet{Seber-and-Braatz-2025-LCEN}, four real and one artificial dataset (summarized in Table \ref{Table_diffMCC_LCEN_classification_dataset_list}) are used to test the performance of the classification LCEN algorithm and of ANNs trained with the diffMCC loss function against previously published and our baselines. In addition to testing the machine-learning performance of LCEN, and unlike in \citet{Seber-and-Braatz-2025-LCEN}, feature selection is performed on the same datasets, and the performance of models trained on features selected by LCEN is also evaluated. These classification datasets can be divided into two categories: binary classification (Section \ref{Section_diffMCC_LCEN_classification_results_binary}) and multiclass classification (Section \ref{Section_diffMCC_LCEN_classification_results_multiclass}). All models tested in this work had their hyperparameters selected by 5-fold cross-validation (CV) and this CV procedure was repeated for 3 different seeds so that the average \(\pm\) standard deviation of results can be reported. The separation between training and testing sets varied depending on the dataset and is detailed in Section \ref{Section_diffMCC_LCEN_classification_appendix_datasets}. A list of the hyperparameters evaluated for each model is available in Section \ref{Section_diffMCC_LCEN_classification_appendix_hyperparameters}.

\section{Results}
For all classification tasks in this work, LCEN and ANNs trained with the diffMCC loss function (MLP-diffMCC) are first evaluated as machine learning algorithms against multiple other methods: logistic regression (LR), logistic regression with an L\(_2\) norm-based penalty (RR) \citep{Tikhonov-1963}, logistic regression with an L\(_1\) norm-based penalty (LASSO) \citep{Santosa-and-Symes-1986,Tibshirani-1996}, logistic regression with an L\(_1\) and L\(_2\) norms-based penalty (EN) \citep{Zou-and-Hastie-2005}, random forest (RF) \citep{Ho-1995}, gradient-boosted decision trees (GBDT) \citep{Friedman-2001}, adaptive boosting (AdaB) \citep{Freund-and-Schapire-1997}, support vector machine with radial-basis functions (SVM) \citep{Boser-etal-1992}, and ANNs trained with the (weighted) cross-entropy loss function (MLP-CE). In addition, to highlight the relevance of each individual part of the LCEN algorithm and demonstrate that LCEN is the optimal algorithm also for classification, ablation tests are performed and reported in Section \ref{Section_diffMCC_LCEN_classification_appendix_ablation} of the Appendix.

Next, to evaluate LCEN's performance as a feature selection algorithm, the features selected by LCEN were used to train all the previously mentioned models, and the results obtained with LCEN-selected features were compared to the results obtained with all features, and with the results obtained with features selected by another published method (such as Mann-Whitney tests or knowledge from domain experts).

\subsection{Binary classification} \label{Section_diffMCC_LCEN_classification_results_binary}
LCEN and the diffMCC loss function are first evaluated on two binary classification tasks. The first task (``Heart Failure Clinical Data'') involves predicting heart failure occurrence as a function of patient age, lifestyle (such as whether the patient is a smoker), and values obtained from blood tests \citep{Ahmad-etal-2017}. The dataset is comprised of information from 299 patients, each with 11 different features. \citet{Ahmad-etal-2017} investigated the data with a Cox model and a nomogram, finding that age, renal dysfunction, blood pressure, ejection fraction, and anemia were considered significant for the purpose of predicting heart failure. Later, \citet{Chicco-and-Jurman-2020} used this dataset to build multiple machine learning models for predicting heart failure (Table 4 of their work). Furthermore, they analyzed the features with multiple statistical tests (Mann-Whitney U test, Shapiro-Wilk test, and chi-squared test) and through their random forest models (changes in prediction accuracy and Gini impurity rankings) to determine that serum creatinine and ejection fraction are the most important features in this dataset (Tables 5--8 of \citet{Chicco-and-Jurman-2020}). Finally, they retrained some machine learning models using only these two features and noted an increase in predictive power (Table 9 of that work). Specifically, they note increases in the positive-class F\(_1\) score and MCC, but decreases in the accuracy and macro F\(_1\) score.

Using the same dataset, we first trained multiple machine learning and deep learning models on this heart failure prediction task. When all features\footnote{Other than the time after second check-up, since it cannot be determined \textit{a priori} and thus it is not predictive.} were used (Table \ref{Table_diffMCC_LCEN_classification_results_heart_failure}, second and third columns), MLP models attained a considerably higher performance than traditional machine learning models. In particular, MLPs trained with the diffMCC loss function achieved the highest metrics, reaching a macro F\(_1\) score of 87.7\% and an MCC of 76.1\%. Among the non-deep learning models, LCEN reached the highest performance in both macro F\(_1\) score and MCC metrics, followed by other nonlinear methods. Furthermore, both the LCEN and the MLP-diffMCC models surpassed the RF model of \citet{Chicco-and-Jurman-2020}.

Next, we trained the same models using only the serum creatinine and ejection fraction features, as per \citet{Chicco-and-Jurman-2020}. When compared against the results obtained when using all features, the performance of linear models is enhanced (Table \ref{Table_diffMCC_LCEN_classification_results_heart_failure}, fourth and fifth columns). However, the performance of nonlinear models is decreased. As nonlinear models were the best-performing models when all features were used, the feature selection of \citet{Chicco-and-Jurman-2020} leads to an overall reduction in predictive performance. Among all models tested, using only the features selected by \citet{Chicco-and-Jurman-2020} leads to an average reduction in the macro F\(_1\) score equal to 1.4\% and in MCC equal to 3.5\%.

Finally, we trained the same models for a third time using only the ejection fraction, age, and serum creatinine, which were the features selected by the LCEN models trained on all features. This feature selection procedure leads to increases in predictive performance for all non-deep learning models (Table \ref{Table_diffMCC_LCEN_classification_results_heart_failure}, sixth and seventh columns), including a greater performance increase for linear models than when using the features selected by \citet{Chicco-and-Jurman-2020}. MLP models remain unchanged in their performance, and MLP-diffMCC models are still the highest-performing models. Moreover, the use of fewer features leads to smaller MLPs being trained, yet these smaller MLPs are as performant as the large MLPs trained with all features. Specifically, the MLPs trained with LCEN-selected features are 78.1\%\(\pm\)11.5\% smaller on average. Among all models tested, the use of only LCEN-selected features leads to an average increase in the macro F\(_1\) score equal to 3.3\% and in MCC equal to 10.0\%.

Statistical tests were done with all results to determine the significance of these performance changes. Paired t-tests comparing the metrics obtained when using all features and those obtained when using LCEN-selected features leads to a p-value equal to 2.8\(\times\)10\(^{-4}\) for the macro F\(_1\) scores and 3.2\(\times\)10\(^{-4}\) for the MCCs. When comparing the metrics obtained when using the features of \citet{Chicco-and-Jurman-2020} vs.\ LCEN-selected features, the p-values equal 2.1\(\times\)10\(^{-7}\) for the macro F\(_1\) scores and 2.2\(\times\)10\(^{-7}\) for the MCCs. These results show that the improvements in performance when using LCEN-selected features are very unlikely to be artifacts.

\begin{table}[h]
    \caption{Average (\(\pm\) standard deviation across 3 CV seeds) metrics for models trained on the ``Heart Failure Clinical Data'' dataset. \citet{Chicco-and-Jurman-2020} features are serum creatinine and ejection fraction. LCEN-selected features are ejection fraction, age, and serum creatinine. The ``RF \citep{Chicco-and-Jurman-2020}'' model was the best-performing model from \citet{Chicco-and-Jurman-2020} both when they used all features and with their selected features. \citet{Chicco-and-Jurman-2020} did not report standard deviations for their results, so they are listed here as ``\(\pm\)?''.}
    \label{Table_diffMCC_LCEN_classification_results_heart_failure}
    \centerline{
    \begin{tabular}{| c | c | c || c | c || c | c |}
    \hline
          & \multicolumn{2}{c ||}{All features} & \multicolumn{2}{c ||}{\citet{Chicco-and-Jurman-2020} features} & \multicolumn{2}{c |}{LCEN-selected features} \\
    Model & F\(_1\) score (\%) & MCC (\%) & F\(_1\) score (\%) & MCC (\%) & F\(_1\) score (\%) & MCC (\%) \\
    \hline
    RF \citep{Chicco-and-Jurman-2020} & 68.2\(\pm\)? & 38.4\(\pm\)? & 21.4\(\pm\)? & 41.8\(\pm\)? & N/A & N/A \\
    \hline
    LR          & 73.8 & 47.9 & 78.0 & 56.5 & 79.2 & 58.5 \\
    LASSO       & 73.8\(\pm\)0.0 & 47.9\(\pm\)0.0 & 78.0\(\pm\)0.0 & 56.5\(\pm\)0.0 & 79.2\(\pm\)0.0 & 58.5\(\pm\)0.0 \\
    RR          & 73.8\(\pm\)0.0 & 47.9\(\pm\)0.0 & 78.0\(\pm\)0.0 & 56.5\(\pm\)0.0 & 79.2\(\pm\)0.0 & 58.5\(\pm\)0.0 \\
    EN          & 73.8\(\pm\)0.0 & 47.9\(\pm\)0.0 & 78.0\(\pm\)0.0 & 56.5\(\pm\)0.0 & 79.2\(\pm\)0.0 & 58.5\(\pm\)0.0 \\
    LCEN        & 79.2\(\pm\)0.0 & 58.5\(\pm\)0.0 & 77.8\(\pm\)1.8 & 55.7\(\pm\)3.7 & \multicolumn{2}{c |}{Equal to ``All features''} \\
    SVM         & 77.2\(\pm\)2.7 & 54.5\(\pm\)5.5 & 69.9\(\pm\)1.4 & 40.1\(\pm\)2.5 & 77.5\(\pm\)2.4 & 55.7\(\pm\)3.8 \\
    RF          & 78.1\(\pm\)1.6 & 57.3\(\pm\)3.2 & 73.9\(\pm\)1.8 & 48.4\(\pm\)3.5 & 78.6\(\pm\)2.6 & 58.5\(\pm\)5.0 \\
    GBDT        & 77.6\(\pm\)3.9 & 56.7\(\pm\)8.6 & 76.5\(\pm\)2.9 & 53.4\(\pm\)6.2 & 80.4\(\pm\)1.0 & 61.6\(\pm\)2.3 \\
    AdaB        & 76.2\(\pm\)4.2 & 53.2\(\pm\)7.7 & 73.9\(\pm\)0.2 & 48.1\(\pm\)0.2  & 79.2\(\pm\)1.0 & 60.2\(\pm\)4.1 \\
    MLP-CE      & 84.7\(\pm\)3.0 & 69.3\(\pm\)6.3 & 81.8\(\pm\)1.0 & 63.8\(\pm\)2.1 & 84.9\(\pm\)2.9 & 70.1\(\pm\)6.5 \\
    MLP-diffMCC & 87.7\(\pm\)1.9 & 76.1\(\pm\)3.9 & 81.3\(\pm\)0.1 & 62.8\(\pm\)0.4 & 86.5\(\pm\)0.0 & 73.9\(\pm\)0.0 \\
    \hline
    \end{tabular} }
\end{table}

The next task (``Bank Marketing'') involves predicting whether a customer will subscribe to a term deposit (a.k.a.\ time deposit or certificate of deposit) as a function of customer demographic data (such as age, job, and highest education level completed), customer banking and loan information (such as whether a customer has a loan with the bank), time and method of contact, and macroeconomic data (such as unemployment rate, consumer confidence indices, and the Euribor 3-month interest rate) \citep{Moro-etal-2014}. The dataset is comprised of information from 41,188 calls/potential customers. Originally, each had 150 features; however, the public dataset ``does not include all attributes due to privacy concerns'' \citep{Moro-etal-2014}, and thus only 20 features are available for modeling.\footnote{One of these 20 features is the total duration of the call. Similarly to the ``time after second check-up'' feature in the previous dataset, it cannot be determined \textit{a priori} and was therefore excluded from the dataset.} \citet{Moro-etal-2014} considered multiple methods for feature selection, including a regular forward selection method (which selected 7/150 features) and a composite forward selection method that leverages domain knowledge through an interview with a domain expert (which selected 22/150 features) (Table 4 of \citet{Moro-etal-2014}). \citet{Moro-etal-2014} noted that this composite forward selection method with domain knowledge led to the best results.

Using the public dataset, we first built models for this bank marketing task. Some of the 20 features are categorical and were one-hot encoded, and the total duration of the call feature was excluded, leading to a final dataset with 54 features. In this first scenario, MLP-diffMCC models were again the model with the highest performance, but LCEN had a performance that was higher than that of MLPs trained with the CE loss (Table \ref{Table_diffMCC_LCEN_classification_results_bank_marketing}, second and third columns). Most other models had comparable performance, but, surprisingly, linear models tended to outperform nonlinear models, suggesting a linear relationship between the available features and whether a customer subscribes to a term deposit.

\citet{Moro-etal-2014} stated that their composite forward selection method with domain knowledge, which led to their best results, selected 22/150 features. Out of those 22, 8 were available in the public dataset. After one-hot encoding, this feature selection procedure led to a dataset with 18 features (out of 54). Models trained with only these 18 features consistently perform worse than models trained with all features (Table \ref{Table_diffMCC_LCEN_classification_results_bank_marketing}, fourth and fifth columns). Notably, linear models trained with these 18 features had worse performance than nonlinear models trained with these same features (which is in opposition to the results from models trained with all features). Furthermore, these linear models also had a negative MCC, indicating that they have worse predictive performance than randomly guessing whether a customer will subscribe to a term deposit. Among all models tested, the use of only the features selected by the method of \citet{Moro-etal-2014} leads to an average reduction in the macro F\(_1\) score equal to 23.8\% and in MCC equal to 84.9\%. As mentioned previously, not all features from the original dataset of \citet{Moro-etal-2014} are available in the public dataset, which may help explain this drastically reduced performance across all models. However, this decreased performance also highlights the limitations of the composite forward selection method with domain knowledge previously used.

Finally, these models were trained using only features that were selected by the LCEN models trained on all features, which led to a dataset with 20 features (out of 54). All models trained with these 20 LCEN-selected features had better performance than the respective models trained with all features (Table \ref{Table_diffMCC_LCEN_classification_results_bank_marketing}, sixth and seventh columns).\footnote{With the possible exception of GBDTs. These models had bimodal results across different CV seeds, which explains the high standard deviation reported for these models in Table \ref{Table_diffMCC_LCEN_classification_results_bank_marketing}. The use of 11 CV seeds instead of 3 did not impact the bimodality of the results, and led to only small reductions in the standard deviations of the metrics.} Overall, MLP-diffMCC models trained with LCEN-selected features have the best predictive performance. Again, thanks to the reduction in input features, smaller MLP models were able to be trained, leading to MLPs that are 88.4\%\(\pm\)6.1\% smaller on average when compared to the MLPs trained with all features. Among all models tested, the use of only LCEN-selected features leads to an average increase in the macro F\(_1\) score equal to 1.2\% and in MCC equal to 7.5\%. When GBDT models are excluded, the average increase in the macro F\(_1\) score equals 2.0\% and in MCC equals 9.4\%.

Interestingly, all four of the top-4 features for LCEN (in decreasing order of importance: the employment variation rate, the Euribor 3-month interest rate, the consumer price index, and the number of employed citizens) were also selected by the composite forward selection method with domain knowledge of \citet{Moro-etal-2014}. However, the relative importances attached by the method of \citet{Moro-etal-2014} to these features were considerably different: only one of these top-4 features for LCEN, the Euribor 3-month interest rate, was in the top-5 of their method; and only three were in the top-10 of their method. These results suggest limitations in their methodology beyond those caused by the exclusion of features from the public dataset, and highlight the performance of LCEN as a feature selection algorithm.

Again, statistical tests were done with all results to determine the significance of these performance changes. Paired t-tests comparing the metrics obtained when using all features and those obtained when using LCEN-selected features lead to a p-value equal to 0.95 for the macro F\(_1\) scores and 0.52 for the MCCs. When GBDT models are excluded from these tests, the p-values equal 6.2\(\times\)10\(^{-5}\) for the macro F\(_1\) scores and 3.2\(\times\)10\(^{-3}\) for the MCCs. When comparing the metrics obtained when using the features selected by the method of \citet{Moro-etal-2014} vs.\ LCEN-selected features, the p-values equal 6.4\(\times\)10\(^{-12}\) for the macro F\(_1\) scores and 2.0\(\times\)10\(^{-19}\) for the MCCs. These results show that the improvements in performance when using LCEN-selected features are unlikely to be artifacts, particularly when GBDT models are excluded from the comparison.

\begin{table}[h]
    \caption{Average (\(\pm\) standard deviation across 3 CV seeds for most models and 11 for GBDT) metrics for models trained on the ``Bank Marketing'' dataset. \citet{Moro-etal-2014} features include 18/54 features that were selected by \cite{Moro-etal-2014} and were available in the public dataset. LCEN-selected features include 20/54 features from the public dataset.}
    \label{Table_diffMCC_LCEN_classification_results_bank_marketing}
    \centerline{
    \begin{tabular}{| c | c | c || c | c || c | c |}
    \hline
          & \multicolumn{2}{c ||}{All features} & \multicolumn{2}{c ||}{\citet{Moro-etal-2014} features} & \multicolumn{2}{c |}{LCEN-selected features} \\
    Model & F\(_1\) score (\%) & MCC (\%) & F\(_1\) score (\%) & MCC (\%) & F\(_1\) score (\%) & MCC (\%) \\
    \hline
    LR          & 64.7 & 30.9 & 48.6 & -1.91 & 65.2 & 30.5 \\
    LASSO       & 64.5\(\pm\)0.2 & 30.7\(\pm\)0.2 & 48.6\(\pm\)0.0 & -1.91\(\pm\)0.00 & 65.2\(\pm\)0.1 & 30.3\(\pm\)0.1 \\
    RR          & 64.7\(\pm\)0.1 & 31.0\(\pm\)0.1 & 48.6\(\pm\)0.0 & -1.91\(\pm\)0.00 & 65.3\(\pm\)0.0 & 30.6\(\pm\)0.1 \\
    EN          & 64.6\(\pm\)0.2 & 30.9\(\pm\)0.2 & 48.6\(\pm\)0.0 & -1.91\(\pm\)0.00 & 65.2\(\pm\)0.1 & 30.4\(\pm\)0.2 \\
    LCEN        & 65.2\(\pm\)0.0 & 30.4\(\pm\)0.1 & 50.5\(\pm\)1.1 & 2.85\(\pm\)0.62 & \multicolumn{2}{c |}{Equal to ``All features''} \\
    SVM         & 63.0\(\pm\)0.2 & 27.1\(\pm\)0.4 & 41.2\(\pm\)6.0 & 3.16\(\pm\)3.54 & 63.2\(\pm\)2.6 & 29.7\(\pm\)6.5 \\
    RF          & 62.1\(\pm\)1.0 & 24.8\(\pm\)1.9 & 51.8\(\pm\)1.2 & 12.0\(\pm\)1.3  & 65.2\(\pm\)0.5 & 32.8\(\pm\)0.7 \\
    GBDT        & 59.6\(\pm\)5.8 & 23.2\(\pm\)8.1 & 50.3\(\pm\)0.6 & 5.11\(\pm\)5.11 & 56.5\(\pm\)10.4 & 21.4\(\pm\)11.5 \\
    AdaB        & 63.4\(\pm\)0.0 & 26.9\(\pm\)0.0 & 34.3\(\pm\)0.0 & 0.0\(\pm\)0.00 & 65.0\(\pm\)0.0 & 30.4\(\pm\)0.0 \\
    MLP-CE      & 63.2\(\pm\)0.1 & 26.4\(\pm\)0.3 & 55.4\(\pm\)0.6 & 11.3\(\pm\)1.6  & 65.4\(\pm\)0.5 & 30.8\(\pm\)0.8 \\
    MLP-diffMCC & 65.5\(\pm\)0.8 & 31.4\(\pm\)2.1 & 55.1\(\pm\)0.4 & 11.1\(\pm\)1.0  & 66.6\(\pm\)0.2 & 33.9\(\pm\)0.5 \\
    \hline
    \end{tabular} }
\end{table}

\subsection{Investigating the effects of the \texttt{LCEN\_clip\_min\_classes\_selected} hyperparameter} \label{Section_diffMCC_LCEN_classification_results_clip_min_class}
As commented in Section \ref{Section_diffMCC_LCEN_classification_modifying_LCEN}, the LCEN algorithm generates one feature set per class when used in multiclass classification tasks, and these feature sets must be harmonized (through the \texttt{LCEN\_clip\_min\_classes\_selected} hyperparameter in this work). Thus, we investigate the effects of this hyperparameter on LCEN's feature selection capabilities using four different variations of an artificial dataset built with sklearn's \texttt{sklearn.datasets.make\_classification()} function. These are called ``Artificial data'' in this work. These datasets all have 750 samples and 300 features, 200 of which are informative and 100 that are completely random (thus, useless). We create dataset variations by changing the number of classes and the distribution of classes in each dataset, such that the first variant has 3 classes and a 40\%-30\%-30\% class distribution (balanced), the second has 3 classes and a 45\%-45\%-10\% class distribution (imbalanced), the third has 4 classes and a 25\% for all classes distribution (balanced), and the fourth has 4 classes and a 40\%-40\%-10\%-10\% class distribution (imbalanced).

The \texttt{LCEN\_clip\_min\_classes\_selected} hyperparameter can vary between 1 and the number of classes in a dataset, and LCEN models trained with this hyperparameter equal to \(k\) will be labeled as ``LCEN-k'' henceforth in this work. The results of these experiments are provided in Table \ref{Table_diffMCC_LCEN_classification_results_LCEN_clip_FS} and allow two important conclusions to be drawn. First, there is a strong reduction in performance as \texttt{LCEN\_clip\_min\_classes\_selected} approaches the number of classes in a dataset (Table \ref{Table_diffMCC_LCEN_classification_results_LCEN_clip_FS}, last two rows). Furthermore, there is greater performance variation among the different seeds in these scenarios. Based on the results here and in Section \ref{Section_diffMCC_LCEN_classification_results_multiclass}, we recommend setting \texttt{LCEN\_clip\_min\_classes\_selected} to no more than 2/3 of the number of classes. Second, the performance of LCEN as a feature selection algorithm does not appear to be affected by class imbalance (Table \ref{Table_diffMCC_LCEN_classification_results_LCEN_clip_FS}, second row vs. third, and fourth row vs. fifth), although it may increase the instability between different CV seeds under the same conditions. These results have guided the selection of the \texttt{LCEN\_clip\_min\_classes\_selected} hyperparameter in Section \ref{Section_diffMCC_LCEN_classification_results_multiclass} and can also act as a guide for general scenarios.

\begin{table}[h]
    \caption{Average (\(\pm\) standard deviation across 3 CV seeds) feature selection metrics for LCEN models trained on the ``Artificial data'' datasets. The numbers for each variant refers to the class distribution of the samples.}
    \label{Table_diffMCC_LCEN_classification_results_LCEN_clip_FS}
    \centerline{
    \begin{tabular}{| c | c | c || c | c |}
    \hline
          Variant: & \multicolumn{2}{c ||}{40\%-30\%-30\%} & \multicolumn{2}{c |}{45\%-45\%-10\%} \\
    Model & F\(_1\) score (\%) & MCC (\%) & F\(_1\) score (\%) & MCC (\%) \\
    \hline
    LCEN-1 & 50.9\(\pm\)1.8 & 15.8\(\pm\)1.5 & 60.7\(\pm\)1.2 & 22.3\(\pm\)3.0 \\
    LCEN-2 & 56.7\(\pm\)2.0 & 16.0\(\pm\)4.2 & 56.0\(\pm\)2.2 & 17.8\(\pm\)1.7 \\
    LCEN-3 & 47.6\(\pm\)2.3 &  5.4\(\pm\)4.2 & 49.4\(\pm\)8.2 &  8.0\(\pm\)7.8 \\
    \hline
    \end{tabular} }
    \vspace{1ex}
    \centerline{
    \begin{tabular}{| c | c | c || c | c |}
    \hline
          Variant: & \multicolumn{2}{c ||}{25\%-25\%-25\%-25\%} & \multicolumn{2}{c |}{40\%-40\%-10\%-10\%} \\
    Model & F\(_1\) score (\%) & MCC (\%) & F\(_1\) score (\%) & MCC (\%) \\
    \hline
    LCEN-1 & 50.5\(\pm\)3.2 & 13.6\(\pm\)1.3 & 46.2\(\pm\)5.6 & 8.8\(\pm\)2.6 \\
    LCEN-2 & 51.1\(\pm\)3.5 & 12.2\(\pm\)0.5 & 56.2\(\pm\)1.8 & 14.0\(\pm\)1.6 \\
    LCEN-3 & 54.4\(\pm\)0.9 &  9.6\(\pm\)2.4 & 40.1\(\pm\)7.5 & 10.4\(\pm\)5.9 \\
    LCEN-4 & 47.8\(\pm\)4.8 &  8.1\(\pm\)2.7 & 36.6\(\pm\)10.8 & 10.5\(\pm\)0.5 \\
    \hline
    \end{tabular} }
\end{table}

\subsection{Multiclass classification} \label{Section_diffMCC_LCEN_classification_results_multiclass}
Next, LCEN and the diffMCC loss function are evaluated on two multiclass classification datasets. The first multiclass classification dataset (``Wine Quality'') involves predicting the median evaluation given by wine experts to a wine as a function of its physicochemical properties, such as density, pH, and the concentrations of certain chemicals \citep{Cortez-etal-2009}. There are two versions of this dataset: one for red wines and another for white wines; this work uses the red wines version. \citet{Cortez-etal-2009} investigated this task using multiple models, the best of which was an SVM (Tables 2 and 3 of that work). Moreover, \citet{Cortez-etal-2009} performed a sensitivity analysis on the inputs of their SVM models to determine which features had the highest importance (Fig.\ 4 of that work); this approach was used to train their final SVM model, but the actual selected features were not reported.

We slightly modified the dataset to remove 10/1,599 (0.63\%) samples of red wine that received a grade of 3, as this was an exceptionally rare grade. Using this modified dataset, we first trained multiple data-driven models on this wine quality prediction task. As in the binary classification tasks, MLPs trained with the diffMCC loss function achieved the highest test-set performance (Table \ref{Table_diffMCC_LCEN_classification_results_wine}, second and third columns), followed by MLPs trained with the cross-entropy loss and then LCEN. They were followed by other nonlinear models. In particular, RF obtained high MCC metrics on this task. \citet{Cortez-etal-2009} reported only whole-dataset results for their SVM (Table 3 of that work), which stated that their SVM reaches a whole-dataset macro F\(_1\) score equal to 36.7\% and an MCC equal to 39.2\%. On the test-set alone, MLP-diffMCC models surpass these values, and MLP-CE, LCEN, and RF models reach comparable values (Table \ref{Table_diffMCC_LCEN_classification_results_wine}, second and third columns).

In this first test with all features, LCEN-1 models showed the best performance when compared to LCEN models with other values of \texttt{LCEN\_clip\_min\_classes\_selected}. The performance of LCEN models decreased slightly with increasing values of \texttt{LCEN\_clip\_min\_classes\_selected}, but the performance decreased considerably when \texttt{LCEN\_clip\_min\_classes\_selected} = 5 (Table \ref{Table_diffMCC_LCEN_classification_clip_min_classes_wine}), and a few LCEN models in that scenario did not select any features. Tukey's HSD test was used to compare the macro F\(_1\) and MCC metrics for LCEN-1 to LCEN-4. The test showed that the differences in macro F\(_1\) scores are not statistically significant (lowest p = 0.076, for the comparison between LCEN-1 and LCEN-4), but the differences in MCC can be significant when compared against LCEN-4 (p = 1.7\(\times\)10\(^{-3}\) for the comparison between LCEN-1 and LCEN-4, p = 0.012 for LCEN-2 and LCEN-4, and p = 0.026 for LCEN-3 and LCEN-4). This analysis suggests that the \texttt{LCEN\_clip\_min\_classes\_selected} hyperparameter should be no higher than 3 for this dataset, which follows the conclusions drawn in Section \ref{Section_diffMCC_LCEN_classification_results_clip_min_class}.

As mentioned above, \citet{Cortez-etal-2009} performed a sensitivity analysis on the inputs of their SVM and obtained quantitative values for each feature; however, they did not provide any recommendations or cutoffs for feature selection. Because LCEN models frequently included 7 features, we used the 7 most important features according to \citet{Cortez-etal-2009}'s analysis to build another set of features. Models trained with these features performed nearly identically to models trained with all features (Table \ref{Table_diffMCC_LCEN_classification_results_wine}, fourth and fifth columns). No clear patterns emerged among the improvements and decreases in the models’ performance. Across all models tested, using the 7 most important features according to \citet{Cortez-etal-2009} leads to an average reduction in the macro F\(_1\) score equal to 0.2\% and in MCC equal to 1.3\%, showing that the method of \citet{Cortez-etal-2009} is viable for feature selection in this task.

The multiple LCEN models\footnote{Specifically, LCEN-1 and LCEN-2 did not exclude any features, but the other versions did. This suggests that every feature may have a non-negligible importance.} tended to frequently exclude 4/11 features: density, pH, volatile acidity, and residual sugar. Models trained on a dataset that excluded these features again performed very similarly to models trained with all features (Table \ref{Table_diffMCC_LCEN_classification_results_wine}, sixth and seventh columns). In this scenario, linear models had reductions in performance, whereas most nonlinear models had increases. Among these models, MLP-diffMCC models had the highest test-set metrics, but these metrics are similar to those obtained by the MLP-diffMCC models trained with all features. Specifically, MLP-diffMCC models trained with LCEN-selected features had a higher macro F\(_1\) score but a lower MCC (Table \ref{Table_diffMCC_LCEN_classification_results_wine}, last row). Among all models tested, the use of LCEN-selected features leads to an average reduction in the macro F\(_1\) score equal to 2.5\% and in MCC equal to 1.5\%. When only nonlinear models are considered, the use of LCEN-selected features increases the average macro F\(_1\) score by 0.1\% and increases the average MCC by 2.9\%. These results show that LCEN is also viable for feature selection in this task, especially for nonlinear models.

Lastly, statistical tests were done with all results to determine the significance of these performance changes. Paired t-tests comparing the metrics obtained when using all features and those obtained when using LCEN-selected features lead to a p-value equal to 0.025 for the macro F\(_1\) scores and 0.50 for the MCCs. When only nonlinear models are considered, the p-values equal 0.999 for the macro F\(_1\) scores and 0.11 for the MCCs. When comparing the metrics obtained when using the 7 most important features according to \citet{Cortez-etal-2009}'s analysis vs.\ LCEN-selected features, the p-values equal 0.020 for the macro F\(_1\) scores and 0.69 for the MCCs. When only nonlinear models are considered, the p-values equal 0.38 for the macro F\(_1\) score and 9.7\(\times\)10\(^{-3}\) for the MCC. These results confirm that both feature selection methods do not lead to statistically significant changes in performance in this task, and can be viable methods to lower the number of features considered.

\begin{table}[h]
    \caption{Average (\(\pm\) standard deviation across 3 CV seeds) metrics for models trained on the ``Wine Quality'' dataset. \citet{Cortez-etal-2009} features include the top 7 features as identified by \citet{Cortez-etal-2009}'s method (sulphates to fixed acidity, as stated in the top half of Fig.\ 4 of that work). LCEN-selected features include the 7 features that were most frequently selected by LCEN models.}
    \label{Table_diffMCC_LCEN_classification_results_wine}
    \centerline{
    \begin{tabular}{| c | c | c || c | c || c | c |}
    \hline
          & \multicolumn{2}{c ||}{All features} & \multicolumn{2}{c ||}{\citet{Cortez-etal-2009} features} & \multicolumn{2}{c |}{LCEN-selected features} \\
    Model & F\(_1\) score (\%) & MCC (\%) & F\(_1\) score (\%) & MCC (\%) & F\(_1\) score (\%) & MCC (\%) \\
    \hline
    LR          & 30.8 & 26.6 & 31.2 & 26.0 & 27.9 & 23.0 \\
    LASSO       & 30.8\(\pm\)0.0 & 26.5\(\pm\)0.2 & 29.5\(\pm\)1.6 & 25.9\(\pm\)0.2 & 27.6\(\pm\)0.4 & 22.5\(\pm\)0.4 \\
    RR          & 30.8\(\pm\)0.0 & 26.6\(\pm\)0.0 & 30.2\(\pm\)1.8 & 25.8\(\pm\)0.4 & 27.8\(\pm\)0.2 & 22.8\(\pm\)0.5 \\
    EN          & 30.8\(\pm\)0.0 & 26.6\(\pm\)0.0 & 29.2\(\pm\)1.8 & 25.6\(\pm\)0.4 & 27.6\(\pm\)0.4 & 22.5\(\pm\)0.4 \\
    LCEN-1      & 36.3\(\pm\)1.0 & 34.3\(\pm\)1.0 & 34.7\(\pm\)1.5 & 34.6\(\pm\)0.9 & 33.2\(\pm\)0.1 & 33.4\(\pm\)0.7 \\
    SVM         & 35.7\(\pm\)0.3 & 37.4\(\pm\)2.4 & 34.8\(\pm\)2.6 & 36.5\(\pm\)4.6 & 37.3\(\pm\)0.4 & 40.8\(\pm\)0.9 \\
    RF          & 36.1\(\pm\)1.0 & 40.9\(\pm\)2.3 & 35.1\(\pm\)1.1 & 34.9\(\pm\)2.6 & 35.8\(\pm\)0.2 & 39.6\(\pm\)0.2 \\
    GBDT        & 32.6\(\pm\)2.9 & 33.2\(\pm\)5.9 & 35.0\(\pm\)0.9 & 36.1\(\pm\)2.6 & 35.6\(\pm\)1.0 & 39.2\(\pm\)2.3 \\
    AdaB        & 31.2\(\pm\)0.3 & 24.8\(\pm\)0.2 & 31.5\(\pm\)0.6 & 25.7\(\pm\)1.2 & 30.6\(\pm\)0.5 & 25.3\(\pm\)1.0 \\
    MLP-CE      & 36.6\(\pm\)0.6 & 40.9\(\pm\)1.2 & 36.8\(\pm\)0.6 & 39.8\(\pm\)0.3 & 37.3\(\pm\)0.2 & 40.5\(\pm\)0.1 \\
    MLP-diffMCC & 40.0\(\pm\)0.7 & 41.3\(\pm\)0.9 & 41.1\(\pm\)0.6 & 39.1\(\pm\)0.4 & 40.3\(\pm\)1.5 & 40.4\(\pm\)0.7 \\
    \hline
    \end{tabular} }
\end{table}

The final task investigated in this work (``Glass Identification'') involves predicting the type (float-processed, non-float-processed, or non-window glass) of a piece of glass as a function of its refractive index and chemical composition, taking into account the presence of 8 elements \citep{German-1987}. Previous investigations by the same author and others have highlighted the value of chemical composition to identify glass types \citep{Dabbs-etal-1973,Butterworth-etal-1974}, but machine learning methods have not been employed by these authors.\footnote{Likely due to the age of these papers, which predate many machine learning developments.} The original dataset features 6 types of glass that can be grouped into the 3 aforementioned types, and this work uses the version with 3 types.

Using this modified dataset, we built multiple machine learning models for this glass identification task. In this first scenario, in which all features were used, MLPs trained with the diffMCC loss function again achieved the highest overall performance, and were followed by MLP-CE, RF, and LCEN models (Table \ref{Table_diffMCC_LCEN_classification_results_glass}, second and third columns). As in most other tasks investigated in this work, these models were followed by other nonlinear models, then by linear models.

LCEN-1 models again performed better than models with other values of the \texttt{LCEN\_clip\_min\_classes\_selected} hyperparameter. The performance of LCEN models again decreased slightly with increasing values of \texttt{LCEN\_clip\_min\_classes\_selected} (Table \ref{Table_diffMCC_LCEN_classification_clip_min_classes_glass}). Unlike what happened with the ``Wine Quality'' task, there was no pronounced performance decrease once \texttt{LCEN\_clip\_min\_classes\_selected} = 3, the number of classes. Tukey's HSD test was used to compare the macro F\(_1\) and MCC metrics for LCEN-1, LCEN-2, and LCEN-3. The test confirmed that the performance differences were not statistically significant, with the lowest p-value for macro F\(_1\) score equal to 0.10 and the lowest p-value for MCC equal to 0.192, both for comparisons between the performance of LCEN-1 and LCEN-3. Nevertheless, we still consider that the conclusions from Section \ref{Section_diffMCC_LCEN_classification_results_clip_min_class} should be followed, and that \texttt{LCEN\_clip\_min\_classes\_selected} should be set to 1 or 2 in tasks with 3 classes.

Although no quantitative feature importance analysis or feature selection was done by \citet{German-1987}, investigations using the correlation of each feature with the classes have been done.\footnote{Available at \href{https://github.com/MachineLearningBCAM/Datasets/blob/main/descr/multi_class_datasets/glass.rst}{github.com/MachineLearningBCAM/Datasets/blob/main/descr/multi\_class\_datasets/glass.rst}.} This analysis suggested that the Mg, Al, Ba, Na, and Fe contents of a glass are the most important features (in order of decreasing importance), and are all more important than its refractive index. Because LCEN models frequently included 4 features, we used the 4 most important features according to this class correlation analysis to build another feature set. Linear models trained with these features had increased performance, whereas most nonlinear models (other than SVM) had decreased performance (Table \ref{Table_diffMCC_LCEN_classification_results_glass}, fourth and fifth columns).

LCEN models tended to primarily select 4 features: Mg, Al, K, and Ca (in order of decreasing importance). Mg and Al were also the two most important features according to the class correlation analysis above, but K and Ca were the two \textit{least} important according to that analysis. Models trained with these features consistently showed increased performance over models trained with all features or with the features selected by class correlation (Table \ref{Table_diffMCC_LCEN_classification_results_glass}, sixth and seventh columns). As with the other datasets, MLP-diffMCC models attained the highest performance also with LCEN-selected features, reaching a test-set average macro F\(_1\) score equal to 90.0\% and an average MCC equal to 84.6\%. Among all models tested, the use of LCEN-selected features leads to an average increase in the macro F\(_1\) score equal to 5.4\%, and an average increase in the MCC equal to 19.1\%.

Finally, statistical tests were again done with all results to determine the significance of these performance changes. Paired t-tests comparing the metrics obtained when using all features and those obtained when using LCEN-selected features lead to a p-value equal to 0.024 for the macro F\(_1\) scores and 2.7\(\times\)10\(^{-5}\) for the MCCs. When comparing the metrics obtained when using the correlation of the features with the classes vs.\ LCEN-selected features, the p-values equal 2.1\(\times\)10\(^{-7}\) for the macro F\(_1\) scores and 2.2\(\times\)10\(^{-7}\) for the MCCs. These results show that the improvements in performance from LCEN-selected features are unlikely to be artifacts.

\begin{table}[h]
    \caption{Average (\(\pm\) standard deviation across 3 CV seeds) metrics for models trained on the ``Glass Identification'' dataset. \citet{Cortez-etal-2009} features include the top 7 features as identified by \citet{Cortez-etal-2009}'s method (sulphates to fixed acidity, as stated in the top half of Fig.\ 4 of that work). LCEN-selected features include the 7 features that were most frequently selected by LCEN models.}
    \label{Table_diffMCC_LCEN_classification_results_glass}
    \centerline{
    \begin{tabular}{| c | c | c || c | c || c | c |}
    \hline
          & \multicolumn{2}{c ||}{All features} & \multicolumn{2}{c ||}{Class correlation features} & \multicolumn{2}{c |}{LCEN-selected features} \\
    Model & F\(_1\) score (\%) & MCC (\%) & F\(_1\) score (\%) & MCC (\%) & F\(_1\) score (\%) & MCC (\%) \\
    \hline
    LR          & 63.0 & 39.8 & 72.3 & 57.6 & 75.2 & 60.8 \\
    LASSO       & 62.6\(\pm\)3.8 & 39.9\(\pm\)6.7 & 72.3\(\pm\)0.0 & 57.6\(\pm\)0.0 & 75.2\(\pm\)0.0 & 60.8\(\pm\)0.0 \\
    RR          & 62.3\(\pm\)3.4 & 39.1\(\pm\)5.6 & 72.3\(\pm\)0.0 & 57.6\(\pm\)0.0 & 75.2\(\pm\)0.0 & 60.8\(\pm\)0.0 \\
    EN          & 64.1\(\pm\)5.1 & 42.2\(\pm\)8.6 & 72.3\(\pm\)0.0 & 57.6\(\pm\)0.0 & 75.2\(\pm\)0.0 & 60.8\(\pm\)0.0 \\
    LCEN-1      & 80.7\(\pm\)2.6 & 69.3\(\pm\)4.7 & 76.3\(\pm\)1.4 & 63.8\(\pm\)2.3 & 83.4\(\pm\)2.6 & 75.0\(\pm\)3.7 \\
    SVM         & 74.9\(\pm\)1.4 & 58.7\(\pm\)2.3 & 79.7\(\pm\)2.0 & 69.3\(\pm\)3.4 & 78.4\(\pm\)7.2 & 69.8\(\pm\)9.8 \\
    RF          & 81.9\(\pm\)1.0 & 72.7\(\pm\)2.0 & 78.2\(\pm\)2.6 & 65.9\(\pm\)4.0 & 86.9\(\pm\)2.7 & 79.9\(\pm\)4.0 \\
    GBDT        & 77.7\(\pm\)7.1 & 65.4\(\pm\)11.0 & 76.4\(\pm\)6.8 & 64.4\(\pm\)11.2 & 80.1\(\pm\)5.3 & 69.0\(\pm\)9.2 \\
    AdaB        & 70.5\(\pm\)3.2 & 55.8\(\pm\)4.7 & 53.7\(\pm\)0.7 & 50.8\(\pm\)7.2 & 53.7\(\pm\)0.7 & 50.8\(\pm\)7.2 \\
    MLP-CE      & 87.0\(\pm\)0.4 & 79.5\(\pm\)0.3 & 83.2\(\pm\)1.3 & 74.5\(\pm\)1.8 & 88.7\(\pm\)0.5 & 82.4\(\pm\)0.3 \\
    MLP-diffMCC & 89.5\(\pm\)0.3 & 83.4\(\pm\)0.5 & 83.0\(\pm\)3.1 & 73.0\(\pm\)4.1 & 90.0\(\pm\)1.6 & 84.6\(\pm\)2.2 \\
    \hline
    \end{tabular} }
\end{table}

\section{Discussion} \label{Section_diffMCC_LCEN_classification_discussion}

This work modifies the LASSO-Clip-EN (LCEN) algorithm of \citet{Seber-and-Braatz-2025-LCEN} to also work with classification tasks (Section \ref{Section_diffMCC_LCEN_classification_modifying_LCEN}). The performance of this modified LCEN algorithm, for both feature selection and machine learning, and of MLPs trained with the diffMCC loss function of \cite{Seber-2024} was systematically evaluated using four different classification datasets (Sections \ref{Section_diffMCC_LCEN_classification_results_binary} and \ref{Section_diffMCC_LCEN_classification_results_multiclass}). The effect of LCEN hyperparameters that were introduced when the algorithm was modified for classification was also investigated (Section \ref{Section_diffMCC_LCEN_classification_results_clip_min_class}).

These experiments have validated the applicability of the modified LCEN algorithm in multiple complex classification tasks, including binary and multiclass problems, and tasks related to the physical, biological, and social sciences. As a machine learning algorithm, LCEN had consistently high performance in binary (Tables \ref{Table_diffMCC_LCEN_classification_results_heart_failure} and \ref{Table_diffMCC_LCEN_classification_results_bank_marketing}) and multiclass (Tables \ref{Table_diffMCC_LCEN_classification_results_wine} and \ref{Table_diffMCC_LCEN_classification_results_glass}) classification tasks. LCEN had a test-set performance that was, on average, only 7.3\% smaller in terms of macro F\(_1\) and 15.0\% smaller in terms of MCC than the best model for each task, and LCEN was able to surpass MLP models trained with the weighted CE loss in the ``Bank marketing'' task. As a feature selection algorithm, LCEN consistently created sparse models, eliminating an average of 56\% of all input features and more than 50\% of the input features on 3/4 of the tested datasets. Additionally, the use of LCEN-selected features to train other models led to improvements in their performance in 3/4 of the datasets tested, and unchanged performance in the remaining one. LCEN was better at feature selection than multiple other methods in the tested datasets, including statistical tests such as Mann-Whitney U tests, forward selection with domain knowledge, and class correlation. These properties and strong performance do not limit LCEN's interpretability in any way whatsoever. As written by \citet{Seber-and-Braatz-2025-LCEN}, ``This combination of accuracy and interpretability is essential for the deployment of machine-learning models in performance-critical scenarios, from aviation to medicine.'' Other benefits of LCEN and its limitations have been further discussed in the second paragraph of the discussion section of \citet{Seber-and-Braatz-2025-LCEN} and are also applicable to this variant LCEN algorithm for classification tasks.

In addition to LCEN, these experiments have validated the applicability of deep learning models trained with the diffMCC loss functions in the same multiple complex classification tasks. The diffMCC loss function was used previously in tasks of biological relevance \citep{Seber-2024,Seber-and-Braatz-2025}, but whether its performance improvements could generalize to other tasks had not been determined. In the four datasets tested in this work, MLPs trained with the diffMCC loss function always attained the highest test-set macro F\(_1\) scores and MCCs. In particular, when compared to MLPs trained with the weighted CE loss, MLPs trained with the diffMCC loss always had higher macro F\(_1\) scores and MCCs on the tested datasets. On average, this increase equaled 4.9\% for the macro F\(_1\) score and 8.5\% for the MCC. These results demonstrate that the diffMCC loss is widely applicable to and leads to improved models for classification tasks, both binary and multiclass, beyond those related to biology. A limitation of the experiments in this work is that all used tabular input data, so any performance improvements of the diffMCC loss in text- or image-based tasks have yet to be evaluated.

There are at least two clear future directions for this work, one for LCEN and one for the diffMCC loss. The future direction for LCEN has been identified by \citet{Seber-and-Braatz-2025-LCEN} and ``involves applying the LCEN algorithm to automatically generate physical equations for hybrid model architectures (such as physics-constrained or physics-guided ML), which have high potential for scientific applications \citep{Peng-etal-2021,Willard-etal-2022}.'' The great performance improvements achieved by models using LCEN-selected features solidify this future direction, which would provide a more quantitative impact to LCEN's feature selection capabilities. The future direction for the diffMCC loss involves evaluating models trained (or finetuned) with it on a wider variety of tasks, particularly those that do not use tabular data as input, such as computer vision tasks.

\clearpage
\bibliography{References}
\bibliographystyle{tmlr}
\clearpage
\setcounter{section}{0}
\renewcommand{\thesection}{A\arabic{section}} 
\renewcommand{\thesubsection}{A\arabic{section}.\arabic{subsection}} 
\makeatletter
\setcounter{figure}{0}
\makeatletter 
\renewcommand{\thefigure}{A\@arabic\c@figure}
\makeatother
\setcounter{table}{0}
\makeatletter 
\renewcommand{\thetable}{A\@arabic\c@table}
\makeatother

\section{Description of datasets used in this work} \label{Section_diffMCC_LCEN_classification_appendix_datasets}

Two types of classification tasks are done in this work: binary classification [``Heart Failure Clinical Records'' and ``Bank Marketing'', Section \ref{Section_diffMCC_LCEN_classification_results_binary}] and multiclass classification [``Wine Quality'' and ``Glass Identification'', Section \ref{Section_diffMCC_LCEN_classification_results_multiclass}] (Table \ref{Table_diffMCC_LCEN_classification_dataset_list}). Additionally, one other data set [``Artificial data''] is used to investigate some properties of LCEN in Section \ref{Section_diffMCC_LCEN_classification_results_clip_min_class}.

All models tested in this work had their hyperparameters selected by 5-fold cross-validation. The separation between training and testing sets varied depending on the dataset. For the ``Heart Failure Clinical Records'' dataset, 20\% of the dataset was randomly separated to form the test set (as per \citet{Chicco-and-Jurman-2020}). For the ``Bank Marketing'' dataset, the last 2,058 entries, corresponding to customers contacted in the last year available in the dataset, were used as the test set (as per \citet{Moro-etal-2014}). For the ``Wine Quality'' dataset, 20\% of the dataset was randomly separated to form the test set. For the ``Glass Identification'' dataset, 20\% of the dataset was randomly separated (in a class-stratified way) to form the test set. 

\begin{table}[h]
    \caption{Datasets used in this work and their sources. The binary classification datasets are used in Section \ref{Section_diffMCC_LCEN_classification_results_binary}, ``Artificial data'' is used in Section \ref{Section_diffMCC_LCEN_classification_results_clip_min_class}, and the multiclass classification datasets are used in Section \ref{Section_diffMCC_LCEN_classification_results_multiclass}.}
    \label{Table_diffMCC_LCEN_classification_dataset_list}
    \centerline{
    \begin{tabular}{| c | c |}
    \hline
    Dataset Name & Source \\
    \hline
    Heart Failure Clinical Records & \citet{Ahmad-etal-2017} \href{https://archive.ics.uci.edu/dataset/519/heart+failure+clinical+records}{[link to dataset]} \\
    Bank Marketing                 & \citet{Moro-etal-2014} \href{https://archive.ics.uci.edu/dataset/222/bank+marketing}{[link to dataset]} \\
    \hline
    Artificial data & Artificial data generated by us \\
    \hline
    Wine Quality & \citet{Cortez-etal-2009} \href{https://archive.ics.uci.edu/dataset/186/wine+quality}{[link to dataset]} \\
    Glass Identification & \citet{German-1987} \href{https://archive.ics.uci.edu/dataset/42/glass+identification}{[link to dataset]} \\
    \hline
    \end{tabular} }
\end{table}

\section{List of hyperparameters used in this work} \label{Section_diffMCC_LCEN_classification_appendix_hyperparameters}
All possible permutations of the hyperparameters below were cross-validated.
\begin{enumerate}
    \item For the LASSO and Ridge regression models: \(\alpha = 0\) and 20 log-spaced values between \(-4.3\) and 0 (as per \texttt{np.logspace(-4.3,0,20)}).
    \item For the elastic net (EN) models: \(\alpha\) as above and L1 ratios equal to [0, 0.1, 0.2, 0.3, 0.4, 0.5, 0.6, 0.7, 0.8, 0.9, 0.95, 0.97, 0.99].
    \item For the LCEN models: \(\alpha\) and L1 ratios as above. \textit{degree} values equal to [1, 2, 3] and \textit{lag} = 0 were used. \textit{cutoff} values between $1$$\times$$10^{-2}$ and $1$$\times$$10^{-1}$ were used for most datasets, and values between $1$$\times$$10^{-2}$ and $6$$\times$$10^{-1}$ were used for the ``Glass Identification'' dataset. A \textit{cutoff} = 0 is used in the ablation tests for the LASSO-EN model (Section \ref{Section_diffMCC_LCEN_classification_appendix_ablation}).
    \item For the support vector machine (SVM) models: C values equal to [0.01, 0.1, 1, 10, 50, 100] and gamma values equal to [1/50, 1/10, 1/5, 1/2, 1, 2, 5, 10, 50] divided by the number of features in a dataset were used.
    \item For the random forest (RF) and gradient-boosted decision tree (GBDT) models: [10, 25, 50, 100, 200, 300] trees, maximum tree depth equal to [2, 3, 5, 10, 15, 20, 40], minimum fraction of samples per leaf equal to [0.01, 0.02, 0.05, 0.1], and minimum fraction of samples per tree equal to [0.1, 0.25, 0.333, 0.5, 0.667, 0.75, 1.0]. For the GBDT models, learning rates equal to [0.01, 0.05, 0.1, 0.2] were also used.
    \item For the AdaBoost (AdaB) models: [10, 25, 50, 100, 200, 300] trees/estimators and learning rates equal to [0.01, 0.05, 0.1, 0.2] were used.
    \item For the multilayer perceptron (MLP) models: the hidden layer sizes varied for each dataset and were a function of the number of features F. Representing an MLP with one hidden layer as [X], one with two hidden layers as [X, Y], and one with three hidden layers as [X, Y ,Z], MLPs with hidden layer sizes of \{[6F, 6F, 6F], [6F, 6F, 5F], [6F, 6F, 4F], [6F, 6F, 3F], [6F, 6F, 2F], [6F, 5F, 5F], [6F, 5F, 4F], [6F, 5F, 3F], [6F, 5F, 2F], [6F, 4F, 4F], [6F, 4F, 3F], [6F, 4F, 2F], [6F, 3F, 3F], [6F, 3F, 2F], [6F, 2F, 2F], [5F, 5F, 5F], [5F, 5F, 4F], [5F, 5F, 3F], [5F, 5F, 2F], [5F, 4F, 4F], [5F, 4F, 3F], [5F, 4F, 2F], [5F, 3F, 3F], [5F, 3F, 2F], [5F, 2F, 2F], [4F, 4F, 4F], [4F, 4F, 3F], [4F, 4F, 2F], [4F, 3F, 3F], [4F, 3F, 2F], [4F, 2F, 2F], [3F, 3F, 3F], [3F, 3F, 2F], [3F, 2F, 2F], [2F, 2F, 2F], [6F, 6F], [6F, 5F], [6F, 4F], [6F, 3F], [6F, 2F], [5F, 5F], [5F, 4F], [5F, 3F], [5F, 2F], [5F, F], [4F, 4F], [4F, 3F], [4F, 2F], [4F, F], [3F, 3F], [3F, 2F], [3F, F], [2F, 2F], [2F, F], [F, F], [F, F/2], [6F], [5F], [4F], [3F], [2F], [F], [F/2]\} were used. Learning rates equal to [0.0005, 0.001, 0.005, 0.01, 0.05], the AdamW optimizer, the ReLU and tanhshrink activation functions, batch sizes between 32--2048, weight decay with \(\lambda\) equal to [0, 0.01], 50--100 epochs, and a cosine scheduler with a minimum learning rate equal to 1/16 of the original learning rate with 10 epochs of warm-up were also used. For the MLPs trained with the diffMCC loss functions, \(\gamma\) equal to [1.0, 1.5, 2.0] were also used.

    Class weights varied depending on the task. For the ``Heart Failure Clinical Records'' and ``Bank Marketing'' datasets, class weights equal to \{[1, 1], [1, 2]\} were used. For the ``Wine Quality'' dataset, class weights equal to \{[1, 1, 1, 1, 1], [1, 1, 1, 1, 2], [2, 1, 1, 1, 2]\} were used. For the ``Glass Identification'' dataset, class weights equal to \{[1, 1, 1], [1, 1, 2]\} were used.
\end{enumerate}

\section{Ablation tests} \label{Section_diffMCC_LCEN_classification_appendix_ablation}
As in \cite{Seber-and-Braatz-2025-LCEN}, to highlight the relevance of each individual part of the algorithm and demonstrate that LCEN is the best variation of this algorithm, ablation tests are performed. Three ablated algorithms [LASSO-Clip (LC, the thresholded LASSO), EN-Clip (ENC, the thresholded EN), and LASSO-EN (LEN)] are compared with the original LCEN algorithm. Two variant algorithms, LASSO-Clip-LASSO (LCL) and EN-Clip-EN (ENCEN), are also compared. 

In all datasets investigated, LCEN reached the highest macro F\(_1\) score and MCC metrics (tied in one case with LC, and by itself in all others) (Tables \ref{Table_diffMCC_LCEN_classification_ablation_heart_failure}--\ref{Table_diffMCC_LCEN_classification_ablation_glass}). Unlike in \citet{Seber-and-Braatz-2025-LCEN}, no specific patterns arose among the results from ablated and variant models. For example, ENC tended to have lower performances, and LC was the worst model on one dataset and tied with LCEN for best model on another. Nevertheless, LCEN was the only model that displayed consistently high performance in these experiments, corroborating the notion that it is the best model among all variants and ablations.

Although runtimes were not formally measured, the same pattern reported in \citet{Seber-and-Braatz-2025-LCEN} is present here: models that start with EN are much slower than models that start with the LASSO.

\begin{table}[H]
    \caption{Results of different ablated and variant LCEN algorithms for the ``Heart Failure Clinical Records'' dataset. Compare with Table \ref{Table_diffMCC_LCEN_classification_results_heart_failure}.}
    \label{Table_diffMCC_LCEN_classification_ablation_heart_failure}
    \centerline{
    \begin{tabular}{| c | c | c |}
    \hline
          & \multicolumn{2}{c |}{All features} \\
    Model & F\(_1\) score (\%) & MCC (\%) \\
    \hline
    LC    & 74.4\(\pm\)1.8 & 48.9\(\pm\)3.8 \\
    ENC   & 73.6\(\pm\)3.7 & 47.3\(\pm\)7.3 \\
    LEN   & 77.1\(\pm\)1.9 & 55.6\(\pm\)3.2 \\
    LCL   & 73.6\(\pm\)3.9 & 48.1\(\pm\)8.1 \\
    ENCEN & 74.3\(\pm\)3.2 & 48.8\(\pm\)6.2 \\
    \hline
    LCEN  & 79.2\(\pm\)0.0 & 58.5\(\pm\)0.0 \\
    \hline
    \end{tabular} }
\end{table}

\begin{table}[H]
    \caption{Results of different ablated and variant LCEN algorithms for the ``Bank Marketing'' dataset. Compare with Table \ref{Table_diffMCC_LCEN_classification_results_bank_marketing}.}
    \label{Table_diffMCC_LCEN_classification_ablation_bank_marketing}
    \centerline{
    \begin{tabular}{| c | c | c |}
    \hline
          & \multicolumn{2}{c |}{All features} \\
    Model & F\(_1\) score (\%) & MCC (\%) \\
    \hline
    LC    & 65.2\(\pm\)0.2 & 30.5\(\pm\)0.3 \\
    ENC   & 57.2\(\pm\)0.2 & 16.3\(\pm\)0.5 \\
    LEN   & 57.2\(\pm\)0.0 & 16.7\(\pm\)0.0 \\
    LCL   & 61.1\(\pm\)4.3 & 22.3\(\pm\)8.4 \\
    ENCEN & 57.3\(\pm\)0.1 & 16.4\(\pm\)0.8 \\
    \hline
    LCEN  & 65.2\(\pm\)0.0 & 30.4\(\pm\)0.1 \\
    \hline
    \end{tabular} }
\end{table}

\begin{table}[H]
    \caption{Results of different ablated and variant LCEN algorithms for the ``Wine Quality'' dataset. Compare with Table \ref{Table_diffMCC_LCEN_classification_results_wine}.}
    \label{Table_diffMCC_LCEN_classification_ablation_wine}
    \centerline{
    \begin{tabular}{| c | c | c |}
    \hline
          & \multicolumn{2}{c |}{All features} \\
    Model & F\(_1\) score (\%) & MCC (\%) \\
    \hline
    LC-1    & 34.9\(\pm\)0.0 & 32.4\(\pm\)0.0 \\
    ENC-1   & 34.6\(\pm\)0.4 & 31.5\(\pm\)1.2 \\
    LEN-1   & 35.0\(\pm\)1.2 & 32.4\(\pm\)2.6 \\
    LCL-1   & 33.2\(\pm\)1.1 & 30.1\(\pm\)3.0 \\
    ENCEN-1 & 33.7\(\pm\)1.7 & 31.5\(\pm\)1.3 \\
    \hline
    LCEN-1  & 36.3\(\pm\)1.0 & 34.3\(\pm\)1.0 \\
    \hline
    \end{tabular} }
\end{table}

\begin{table}[H]
    \caption{Results of different ablated and variant LCEN algorithms for the ``Glass Identification'' dataset. Compare with Table \ref{Table_diffMCC_LCEN_classification_results_glass}.}
    \label{Table_diffMCC_LCEN_classification_ablation_glass}
    \centerline{
    \begin{tabular}{| c | c | c |}
    \hline
          & \multicolumn{2}{c |}{All features} \\
    Model & F\(_1\) score (\%) & MCC (\%) \\
    \hline
    LC-1    & 70.9\(\pm\)1.6 & 51.6\(\pm\)5.2 \\
    ENC-1   & 71.5\(\pm\)4.6 & 54.6\(\pm\)7.9 \\
    LEN-1   & 73.0\(\pm\)1.1 & 55.6\(\pm\)1.6 \\
    LCL-1   & 73.7\(\pm\)1.5 & 58.6\(\pm\)1.5 \\
    ENCEN-1 & 76.4\(\pm\)3.8 & 63.1\(\pm\)5.5 \\
    \hline
    LCEN-1  & 80.7\(\pm\)2.6 & 69.3\(\pm\)4.7 \\
    \hline
    \end{tabular} }
\end{table}

\section{Additional results for the multiclass tasks as the \texttt{LCEN\_clip\_min\_classes\_selected} hyperparameter varies}
\begin{table}[H]
    \caption{LCEN results with different values of the \texttt{LCEN\_clip\_min\_classes\_selected} hyperparameter for the ``Wine Quality'' dataset. ``N/A'' stands for combinations not tested.}
    \label{Table_diffMCC_LCEN_classification_clip_min_classes_wine}
    \centerline{
    \begin{tabular}{| c | c | c || c | c || c | c |}
    \hline
          & \multicolumn{2}{c ||}{All features} & \multicolumn{2}{c ||}{\citet{Cortez-etal-2009} features} & \multicolumn{2}{c |}{LCEN-selected features} \\
    Model & F\(_1\) score (\%) & MCC (\%) & F\(_1\) score (\%) & MCC (\%) & F\(_1\) score (\%) & MCC (\%) \\
    \hline
    LCEN-1  & 36.3\(\pm\)1.0 & 34.3\(\pm\)1.0 & 34.7\(\pm\)1.5 & 34.6\(\pm\)0.9 & 33.2\(\pm\)0.1 & 33.4\(\pm\)0.7 \\
    LCEN-2  & 35.1\(\pm\)1.3 & 33.2\(\pm\)0.7 & 35.6\(\pm\)1.4 & 33.7\(\pm\)0.8 & 33.3\(\pm\)0.1 & 33.6\(\pm\)0.3 \\
    LCEN-3  & 33.6\(\pm\)0.3 & 32.8\(\pm\)0.2 & 35.6\(\pm\)1.6 & 34.1\(\pm\)0.8 & 33.9\(\pm\)0.6 & 34.6\(\pm\)0.7 \\
    LCEN-4  & 33.4\(\pm\)1.8 & 30.3\(\pm\)1.1 & N/A & N/A & N/A & N/A \\
    LCEN-5  & 22.6\(\pm\)10.4 & 17.1\(\pm\)15.3 & N/A & N/A & N/A & N/A \\
    \hline
    \end{tabular} }
\end{table}

\begin{table}[H]
    \caption{LCEN results with different values of the \texttt{LCEN\_clip\_min\_classes\_selected} hyperparameter for the ``Glass Identification'' dataset. ``N/A'' stands for combinations not tested.}
    \label{Table_diffMCC_LCEN_classification_clip_min_classes_glass}
    \centerline{
    \begin{tabular}{| c | c | c || c | c || c | c |}
    \hline
          & \multicolumn{2}{c ||}{All features} & \multicolumn{2}{c ||}{Class correlation features} & \multicolumn{2}{c |}{LCEN-selected features} \\
    Model & F\(_1\) score (\%) & MCC (\%) & F\(_1\) score (\%) & MCC (\%) & F\(_1\) score (\%) & MCC (\%) \\
    \hline
    LCEN-1  & 80.7\(\pm\)2.6 & 69.3\(\pm\)4.7 & 76.3\(\pm\)1.4 & 63.8\(\pm\)2.3 & 83.4\(\pm\)2.6 & 75.0\(\pm\)3.7 \\
    LCEN-2  & 77.9\(\pm\)3.7 & 64.2\(\pm\)6.4 & N/A & N/A & N/A & N/A \\
    LCEN-3  & 75.0\(\pm\)2.0 & 60.0\(\pm\)5.8 & N/A & N/A & N/A & N/A \\
    \hline
    \end{tabular} }
\end{table}

\section{Computational resources used}
All experiments were done in a personal computer equipped with a 13th Gen Intel\textsuperscript{\textregistered} Core™ i5-13600K CPU, 64 GB of DDR4 RAM, and an NVIDIA GeForce RTX 4090 GPU.

\end{document}